\documentclass[sigconf]{aamas}

\usepackage{balance} % for balancing columns on the final page

\usepackage[T1]{fontenc}
\usepackage{graphicx}
\usepackage{amsmath}
\usepackage{amsfonts}
\usepackage{xcolor}
\usepackage{xspace}
\usepackage{subcaption}
\usepackage{pifont}
\usepackage{makecell}

\usepackage{amsthm}
\usepackage{mathtools}

\usepackage{amsmath}

\makeatletter
\gdef\@copyrightpermission{
  \begin{minipage}{0.2\columnwidth}
   \href{https://creativecommons.org/licenses/by/4.0/}{\includegraphics[width=0.90\textwidth]{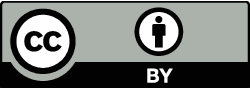}}
  \end{minipage}\hfill
  \begin{minipage}{0.8\columnwidth}
   \href{https://creativecommons.org/licenses/by/4.0/}{This work is licensed under a Creative Commons Attribution International 4.0 License.}
  \end{minipage}
  \vspace{5pt}
}
\makeatother

\setcopyright{ifaamas}
\acmConference[AAMAS '25]{Proc.\@ of the 24th International Conference
on Autonomous Agents and Multiagent Systems (AAMAS 2025)}{May 19 -- 23, 2025}
{Detroit, Michigan, USA}{Y.~Vorobeychik, S.~Das, A.~Nowé  (eds.)}
\copyrightyear{2025}
\acmYear{2025}
\acmDOI{}
\acmPrice{}
\acmISBN{}

\acmSubmissionID{<<submission id>>}

\title[Simulating Tracking Data to Advance Sports Analytics Research]{Simulating Tracking Data to Advance Sports Analytics Research}

\subtitle{Demonstration Track}

\author{David Radke*}
\affiliation{
  \institution{Chicago Blackhawks}
  \city{Chicago}
  \country{USA}}
\email{dradke@blackhawks.com}

\author{Kyle Tilbury*}
\affiliation{
  \institution{University of Waterloo}
  \city{Waterloo}
  \country{Canada}}
\email{ktilbury@uwaterloo.ca}

\begin{abstract}

Advanced analytics have transformed how sports teams operate, particularly in episodic sports like baseball.
Their impact on continuous invasion sports, such as soccer and ice hockey, has been limited due to increased game complexity and restricted access to high-resolution game tracking data.
In this demo, we present a method to collect and utilize simulated soccer tracking data from the Google Research Football environment to support the development of models designed for continuous tracking data.
The data is stored in a schema that is representative of real tracking data and we provide processes that extract high-level features and events.
We include examples of established tracking data models to showcase the efficacy of the simulated data.
We address the scarcity of publicly available tracking data, providing support for research at the intersection of artificial intelligence and sports analytics.
\end{abstract}


\newcommand{\BibTeX}{\rm B\kern-.05em{\sc i\kern-.025em b}\kern-.08em\TeX}

\begin{document}

%%% The following commands remove the headers in your paper. For final 
%%% papers, these will be inserted during the pagination process.

\pagestyle{fancy}
\fancyhead{}

%%% The next command prints the information defined in the preamble.

\maketitle 

\def\thefootnote{*}\footnotetext{These authors contributed equally to this work.}

%%%%%%%%%%%%%%%%%%%%%%%%%%%%%%%%%%%%%%%%%%%%%%%%%%%%%%%%%%%%%%%%%%%%%%%%

\section{Introduction}
\label{sec:intro}

% \begin{itemize}
%     \item Present problem, types of data, types of sports with tracking, etc...
% \end{itemize}

% \todo{Deliverables: 1) collected tracking dataset, 2) script to collect more data if wanted, 3) scripts to make events, models (and analysis)}

% \todo{For demo paper below:}

% \begin{itemize}
%     \item Shrink intro, make it more like related work in one. (DONE? maybe more refs)

%     \item Talk about schema at a high level for tracking tables

%     \item Demo stuff: make events, make models, provide intro scripts for models... Show brief analysis of events/models. Include script to collect more data?

%     \item Discussion: push multiagent approach to sports analytics, this is major step forward
% \end{itemize}

% \todo{Description of paper for demo track: Pages 1-2 should describe the system to be demonstrated. More precisely, the authors are encouraged to discuss the application domain, the problem scenario, the technology used, the agent/multi-agent techniques involved, the system’s innovations, the system’s live and interactive aspects, etc.
% Page 3 should include bibliographic references.
% Page 4 should provide the organisers with a list of requirements for the demo setting at the conference.}

Using advanced analytics to inform decision making in sports has been an increasing trend over recent decades.
% While many major sports leagues across the world have been operational for several decades, the use of data-driven insights and advanced analytics as a basis for roster construction has mostly gained popularity in recent years.
This has been particularly prominent in discrete episodic sports like baseball and has revolutionized how front offices operate~\cite{lewis2004moneyball,elitzur2020data}. 
However, the adoption of analytics in invasion style sports has been relatively limited due to more complex multiagent problems~\cite{radke2023presenting} and limited public access to data~\cite{pleuler2024blog}.

% Specifically, sports in their analytics infancy are those with a variety of multiagent challenges that require further research at the intersection of sports data and multiagent systems~\cite{radke2023presenting}.
% Leading researchers in sports analytics have argued that one of the main limiting factors preventing a flourishing analytics community in invasion sports is public access to tracking data~\cite{pleuler2024blog}.
% While many of these invasion sports showcase a breadth of similarities with existing multiagent research areas, high resolution datasets are hard to come by in the public space.
Most publicly available datasets in invasion sports record \emph{event} data, or play-by-play data, indicating when an on-ball/puck event occurs, players involved, and location; however, event data fails to accurately capture the state of the game during the event (i.e., location of all other players).
At the highest levels, professional ice hockey, basketball, and association football (soccer) leagues collect \emph{tracking} data throughout games, providing the location and velocity of players and the ball/puck multiple times per-second.
Joined with event data, tracking data provides more insight into the state of a game when players choose to take actions (or not to take actions) and has supported research at the intersection of multiagent systems and invasion games, such as measuring player coordination~\cite{spearman2018beyond,radke2022identifying}, improving or evaluating joint actions~\cite{wang2024tacticai,tuyls2021game}, and evaluating individual player actions~\cite{fernandez2021framework}.
% Expanding access to tracking data will support further research at the intersection of multiagent systems and sports~\cite{tuyls2021game,radke2023presenting}.

Despite the crucial potential that tracking data has to support research at understanding inter-player behavior, public access to existing tracking data remains limited.
Tracking datasets are typically protected by bargaining agreements, only giving access to a select few entities involved such as teams, broadcasters, or gambling companies.
As a result, most research at the intersection of artificial intelligence (AI) and sports utilizes event data, limiting the complexity and usability of model architectures.
% Expanding access to tracking data will support further research at the intersection of multiagent systems and sports~\cite{tuyls2021game,radke2023presenting}.

% has been building motivation to address more problems at the intersection of multiagent systems and sports~\cite{tuyls2021game,radke2023presenting}.
% ; further access to tracking data will support research at the intersection of sporting environments and multiagent systems.

In this demo, we present a system to collect simulated ball and player tracking data from the Google Research Football (GRF) reinforcement learning (RL) environment~\cite{kurach2020google} in a way that emulates the schema of real-world ball and player tracking data.
We provide access to a pre-collected dataset that consists of 3,000 simulated soccer games~\cite{dataset} and provide a process to extract event data and different segments of continuous gameplay.
Additionally, our demonstration includes two examples of popular models for player and team evaluation across different invasion sports: expected goals (xG) and pitch control.
We show that the simulated dataset not only is able to support models typically built using real tracking data, but also offers an accessible solution for advancing research in sports analytics and multiagent systems.
% The aim of this work is to support academic and public research in gaining familiarity with tracking data and supporting future models with an emphasis on multiagent considerations.
We include our code and pre-recorded dataset (\href{https://github.com/Dtradke/grf_tracking_data}{\texttt{repository}}) and our demo video (\href{https://youtu.be/2pjyxfPVsuw}{\texttt{video}}).

\section{Tracking Data and Schema}
\label{sec:data_schema}

Most current tracking systems deployed across professional and amateur invasion sports record \emph{center of mass} locations for each player using computer vision or hardware-based systems.
This represents each player and the ball/puck as a three-dimensional coordinate on the playing surface that is referenced using unique player and team identification keys.
While GRF has the ability to represent the game state as a rendered image, the headless state space of the environment represents players and the ball using similar center of mass coordinates.
We utilize the headless representation of the state to record GRF data that imitates the schema of real world invasion game tracking datasets.

We store the recorded tracking data in a schema analogous to real invasion game tracking data~\cite{omidshafiei2022multiagent}.
Each game in our configuration lasts for 3,000 timesteps.
We record center of mass data for all 22 agents on the field and the ball at each timestep, referred to as \emph{entities}.
Thus, each timestep contains 23 consecutive rows of data logging information on each entity for that moment in time, such as identification keys, coordinate locations of the entity in the environment, team affiliation, role on the team, and their current velocity.
Agents are defined with one of eight roles (i.e., positions) within their team, representing different agent \emph{types} that make up a larger team structure.
Agent types and team structures in sporting domains represent similar areas of research across the broader field of multiagent systems~\cite{Radke2022Exploring,albrecht2018autonomous}.
We extract a boolean variable from the environment to indicate if an agent is in possession of the ball.

We include a pre-collected dataset of tracking data from 3,000 simulated GRF games~\cite{dataset}.
Outlined below, our demo provides the ability to collect more tracking data, extract high level features from the raw tracking data, and example models initially published with real football tracking data.

% \section{Demonstration}
\label{sec:demo}

% Our demonstration covers two key components: the process of collecting and processing the data from GRF and the implementation of models utilizing this data.

% \begin{itemize}
%     \item present amount of data, table details, how to make events (and what the schema looks like), detail on each column
% \end{itemize}

% \subsection{Tracking Table}
% \label{subsec:tracking_table}

\section{Collect Data and Extract Events}

\paragraph{Collecting Simulated Data}

Our demo first involves a process to collect simulated data from the GRF environment~\cite{kurach2020google}.
Raw tracking data shows the positions and velocities of all players and the ball at each timestep of the game.

\paragraph{Extracting Events and Stints}

While raw tracking provides a rich view of the state space, various models for invasion game sports involve determining key moments when players make decisions.
We provide a system that extracts higher level \emph{event} information from the raw tracking data such as passes, receptions, shots, turnovers, and interceptions.
Passes and receptions indicate instances where a team transfers possession of the ball from one agent (i.e., player) to a teammate.
Transitions that are not to a teammate are labeled as turnovers and interceptions.
Shots represent events when an agent moves the ball in an attacking direction and either scores a goal, puts the ball out of play beyond the attacking goal line, or transfers possession to the opposing goaltender.
Furthermore, many invasion game models utilize continuous segments of gameplay known as \emph{stints}.
We include a process to automatically identify stints from the raw tracking data and assign them unique identifiers.
Extracting these actionable events and features provides the groundwork for more advanced analyses.

\section{Building Models with the Data}
\label{sec:sample_models}

% \begin{itemize}
%     \item Show xG, pitch control, etc?

%     \item pitch control: https://github.com/Friends-of-Tracking-Data-FoTD/LaurieOnTracking
% \end{itemize}

% we're trying to support public research with tracking data
% we need to check that we can build existing public models with this data
% we include various scripts to recreate simple models (xG, pitch control).

% Public research analyzing invasion game sports seldom utilizes player tracking data due to limited public availability.
% Tracking datasets are typically protected by bargaining and other legal agreements, only giving access to few entities involved in the agreements (i.e., teams, broadcasts, gambling companies).
% For the offline dataset and data collection process provided with this work to have impact in supporting public research, the data must be able to support existing research and models.
% We implement versions of several models using our various data sources and include them in the data collection repository.

\paragraph{Event Data: Expected Goals (xG)}

Expected goals (xG) models aims to estimate the probability that a shot results in a goal.
xG models are common across sports with goaltenders as a way to assign value to both shooters and goaltenders by comparing the predicted probability of a goal with the shot outcome.
These models typically rely on information encoded in event data to model the environment of a shot, such as angle and distance of the shot location.
While more advanced xG models can be developed with tracking data, we include features from our event extraction process to show the feasibility of that process.

\begin{figure}[t]
    \centering
    \begin{subfigure}[b]{0.32\linewidth} 
        \centering
        \includegraphics[width=\linewidth]{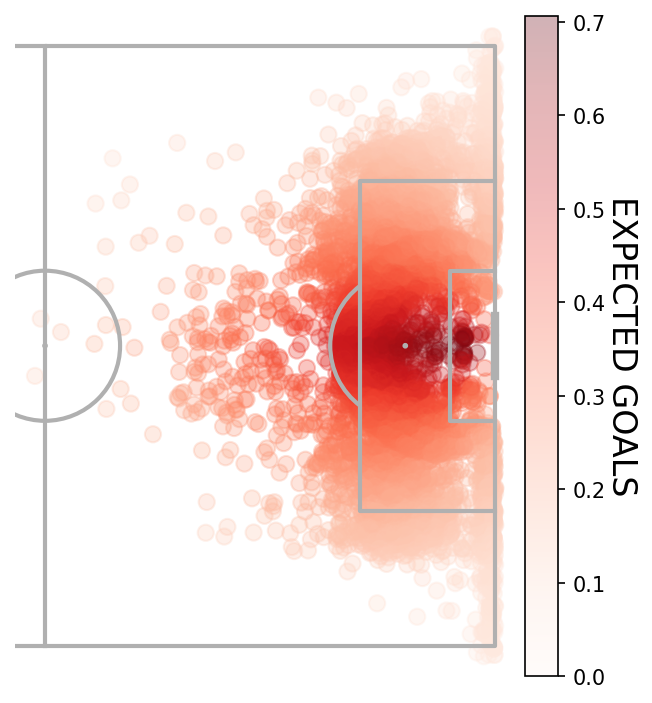}
        \caption{Expected goals.}
        \label{fig:shots_xg}
    \end{subfigure}
    \begin{subfigure}[b]{0.55\linewidth} 
        \includegraphics[width=\linewidth]{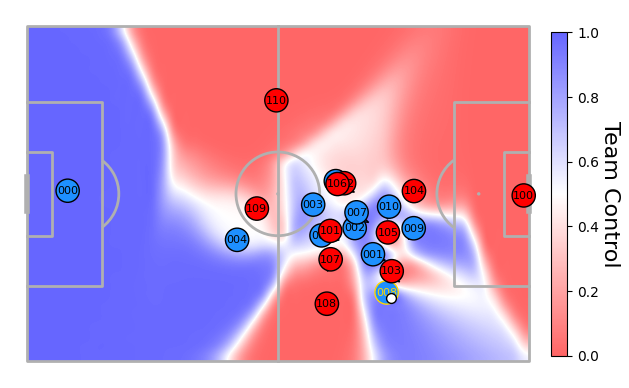}
        \caption{Pitch control.}
        \label{fig:pitch_control}
    \end{subfigure}
        \caption{Examples of expected goals (xG) and pitch control models with our simulated dataset.}
        \label{fig:models}
\end{figure}

We build an xG model using shot angle and distance from the goal.
We align the shot direction so that all shots are modeled as shooting towards the \emph{right} goal (i.e., highest $x$-coordinate direction) and measure shot location as coordinates of where the shot is taken from.
We train a logistic regression model from the set of goals and non-goals using polar coordinates from the center of the opposing goal.
Figure~\ref{fig:shots_xg} shows the results of our logistic regression model on shots extracted from the pre-recorded recorded dataset.
Each dot represents a shot location from either team and darker red represents higher probability of the shot being a goal.
Consistent with work utilizing real soccer data, we observe a monotonic decline in scoring probability as the shot location moves further from the goal at increased angles~\cite{spearman2018beyond}.

% \begin{figure}[t]
% % \begin{subfigure}[b]{0.325\linewidth} 
%     \includegraphics[width=\linewidth]{Figures/shots_xg.png}
%     % \caption{Expected goals (xG) model.}
%     % \label{fig:xg}
% % \end{subfigure}
%     \caption{Results of our expected goals (xG) model.}
%     \label{fig:shots_xg}
% \end{figure}

% \begin{figure*}[t]
%     \centering
%     \begin{subfigure}[b]{0.25\linewidth} 
%         \centering
%         \includegraphics[width=\linewidth]{Figures/shots_goals_half.png}
%         \caption{Successful shots (Goals).}
%         \label{fig:goals}
%     \end{subfigure}
%     \begin{subfigure}[b]{0.25\linewidth} 
%         \includegraphics[width=\linewidth]{Figures/shots_misses_half.png}
%         \caption{Missed shots (Non-goals).}
%         \label{fig:misses}
%     \end{subfigure}
%     \begin{subfigure}[b]{0.325\linewidth} 
%         \includegraphics[width=\linewidth]{Figures/shots_xg.png}
%         \caption{Expected goals (xG) model.}
%         \label{fig:xg}
%     \end{subfigure}
%         \caption{Examples of all shots in our dataset showing (a) goals and (b) non-goals (either misses or saves). Figure~\ref{fig:xg} shows the results of a simple logistic regression expected goals (xG) model built off all shots using angle and direction of each shot to predict the probability of goal.}
%         \label{fig:shots_xg}
% \end{figure*}

\paragraph{Tracking Data: Pitch Control}
\label{subsec:pitch_control}

A common advanced model in invasion sports analytics using tracking data is called \emph{Pitch Control}~\cite{spearman2017physics}.
Given a state of the game represented by tracking data, pitch control aims to quantify the probability that a player would possess the ball if the ball were moved to any location of the pitch.
The union of teammates represents the probability that either team would gain possession of the ball.
The concept of pitch control builds upon traditional Voronoi diagrams~\cite{kim2004voronoi} to include information on player velocity, ball travel speed, and player control time in measuring player and team spatial control over the pitch.
Pitch control models have been used to understand player passing abilities and decisions, including how players take actions to maximize pass completion while also maximizing the pitch \emph{value} with on- and off-ball movements~\cite{spearman2017physics,fernandez2021framework,spearman2018beyond}.

Figure~\ref{fig:pitch_control} shows pitch control for an individual timestep of a simulated game.
Players are divided into blue and red teams and each player has a unique identification number.
The player in possession of the ball (player 008) is highlighted with a gold circle.
Blue regions represent areas where the blue team has a higher probability of recovering the ball if it were moved to those locations and red represents the inverse for the red team.
White locations represent areas where possession probability is close to 50\% for either team.

% \todo{Include figure/analysis? of pitch control model. Could be interesting to show individual pitch control AND team-level pitch control.}
\section{Conclusion}

% \begin{itemize}
%     \item Talk about limitations of the dataset and other limitations in real tracking data.
% \end{itemize}
% This paper presents an offline dataset and data collection process for simulated player and ball tracking data using the Google Research Football environment.
This demonstration presents a data collection and analysis process for simulated player and ball tracking data using the Google Research Football environment.
We show processes to extract additional features from the raw tracking data and show how real tracking-based models can be developed with this simulated data.
Like real tracking data, several limitations exist in this dataset such as the lack of pose estimation, some omitted ball possessions, and various edge cases in the event extraction that are areas for future work.
With the lack of publicly available tracking data for most researchers, this demonstration offers a tangible way to help forward research in artificial intelligence and sports analytics.

%%%%%%%%%%%%%%%%%%%%%%%%%%%%%%%%%%%%%%%%%%%%%%%%%%%%%%%%%%%%%%%%%%%%%%%%

%%% The next two lines define, first, the bibliography style to be 
%%% applied, and, second, the bibliography file to be used.

\bibliographystyle{ACM-Reference-Format} 
\bibliography{main}

%%% -*-BibTeX-*-
%%% Do NOT edit. File created by BibTeX with style
%%% ACM-Reference-Format-Journals [18-Jan-2012].

\begin{thebibliography}{16}

%%% ====================================================================
%%% NOTE TO THE USER: you can override these defaults by providing
%%% customized versions of any of these macros before the \bibliography
%%% command.  Each of them MUST provide its own final punctuation,
%%% except for \shownote{}, \showDOI{}, and \showURL{}.  The latter two
%%% do not use final punctuation, in order to avoid confusing it with
%%% the Web address.
%%%
%%% To suppress output of a particular field, define its macro to expand
%%% to an empty string, or better, \unskip, like this:
%%%
%%% \newcommand{\showDOI}[1]{\unskip}   % LaTeX syntax
%%%
%%% \def \showDOI #1{\unskip}           % plain TeX syntax
%%%
%%% ====================================================================

\ifx \showCODEN    \undefined \def \showCODEN     #1{\unskip}     \fi
\ifx \showDOI      \undefined \def \showDOI       #1{#1}\fi
\ifx \showISBNx    \undefined \def \showISBNx     #1{\unskip}     \fi
\ifx \showISBNxiii \undefined \def \showISBNxiii  #1{\unskip}     \fi
\ifx \showISSN     \undefined \def \showISSN      #1{\unskip}     \fi
\ifx \showLCCN     \undefined \def \showLCCN      #1{\unskip}     \fi
\ifx \shownote     \undefined \def \shownote      #1{#1}          \fi
\ifx \showarticletitle \undefined \def \showarticletitle #1{#1}   \fi
\ifx \showURL      \undefined \def \showURL       {\relax}        \fi
% The following commands are used for tagged output and should be
% invisible to TeX
\providecommand\bibfield[2]{#2}
\providecommand\bibinfo[2]{#2}
\providecommand\natexlab[1]{#1}
\providecommand\showeprint[2][]{arXiv:#2}

\bibitem[\protect\citeauthoryear{Albrecht and Stone}{Albrecht and Stone}{2018}]%
        {albrecht2018autonomous}
\bibfield{author}{\bibinfo{person}{Stefano~V Albrecht} {and} \bibinfo{person}{Peter Stone}.} \bibinfo{year}{2018}\natexlab{}.
\newblock \showarticletitle{Autonomous agents modelling other agents: A comprehensive survey and open problems}.
\newblock \bibinfo{journal}{\emph{Artificial Intelligence}}  \bibinfo{volume}{258} (\bibinfo{year}{2018}), \bibinfo{pages}{66--95}.
\newblock


\bibitem[\protect\citeauthoryear{Elitzur}{Elitzur}{2020}]%
        {elitzur2020data}
\bibfield{author}{\bibinfo{person}{Ramy Elitzur}.} \bibinfo{year}{2020}\natexlab{}.
\newblock \showarticletitle{Data analytics effects in major league baseball}.
\newblock \bibinfo{journal}{\emph{Omega}}  \bibinfo{volume}{90} (\bibinfo{year}{2020}), \bibinfo{pages}{102001}.
\newblock


\bibitem[\protect\citeauthoryear{Fern{\'a}ndez, Bornn, and Cervone}{Fern{\'a}ndez et~al\mbox{.}}{2021}]%
        {fernandez2021framework}
\bibfield{author}{\bibinfo{person}{Javier Fern{\'a}ndez}, \bibinfo{person}{Luke Bornn}, {and} \bibinfo{person}{Daniel Cervone}.} \bibinfo{year}{2021}\natexlab{}.
\newblock \showarticletitle{A framework for the fine-grained evaluation of the instantaneous expected value of soccer possessions}.
\newblock \bibinfo{journal}{\emph{Machine Learning}} \bibinfo{volume}{110}, \bibinfo{number}{6} (\bibinfo{year}{2021}), \bibinfo{pages}{1389--1427}.
\newblock


\bibitem[\protect\citeauthoryear{Kim}{Kim}{2004}]%
        {kim2004voronoi}
\bibfield{author}{\bibinfo{person}{S Kim}.} \bibinfo{year}{2004}\natexlab{}.
\newblock \showarticletitle{Voronoi analysis of a soccer game}.
\newblock \bibinfo{journal}{\emph{Nonlinear Analysis: Modelling and Control}} \bibinfo{volume}{9}, \bibinfo{number}{3} (\bibinfo{year}{2004}), \bibinfo{pages}{233--240}.
\newblock


\bibitem[\protect\citeauthoryear{Kurach, Raichuk, Sta{\'n}czyk, Zaj{\k{a}}c, Bachem, Espeholt, Riquelme, Vincent, Michalski, Bousquet, et~al\mbox{.}}{Kurach et~al\mbox{.}}{2020}]%
        {kurach2020google}
\bibfield{author}{\bibinfo{person}{Karol Kurach}, \bibinfo{person}{Anton Raichuk}, \bibinfo{person}{Piotr Sta{\'n}czyk}, \bibinfo{person}{Micha{\l} Zaj{\k{a}}c}, \bibinfo{person}{Olivier Bachem}, \bibinfo{person}{Lasse Espeholt}, \bibinfo{person}{Carlos Riquelme}, \bibinfo{person}{Damien Vincent}, \bibinfo{person}{Marcin Michalski}, \bibinfo{person}{Olivier Bousquet}, {et~al\mbox{.}}} \bibinfo{year}{2020}\natexlab{}.
\newblock \showarticletitle{Google research football: A novel reinforcement learning environment}. In \bibinfo{booktitle}{\emph{Proceedings of the AAAI conference on artificial intelligence}}, Vol.~\bibinfo{volume}{34}. \bibinfo{pages}{4501--4510}.
\newblock


\bibitem[\protect\citeauthoryear{{Kyle Tilbury, David Radke}}{{Kyle Tilbury, David Radke}}{2024}]%
        {dataset}
\bibfield{author}{\bibinfo{person}{{Kyle Tilbury, David Radke}}.} \bibinfo{year}{2024}\natexlab{}.
\newblock \bibinfo{title}{{Simulated Football Dataset of 3,000 Games}}.
\newblock \bibinfo{howpublished}{\url{https://drive.google.com/drive/folders/1PZ8b-ftnqhIqMV0qnkTB_LWtCnBjhTdD}}.
\newblock
\newblock
\shownote{Accessed 2024-12-17.}


\bibitem[\protect\citeauthoryear{Lewis}{Lewis}{2004}]%
        {lewis2004moneyball}
\bibfield{author}{\bibinfo{person}{Michael Lewis}.} \bibinfo{year}{2004}\natexlab{}.
\newblock \bibinfo{booktitle}{\emph{Moneyball: The art of winning an unfair game}}.
\newblock \bibinfo{publisher}{WW Norton \& Company}.
\newblock


\bibitem[\protect\citeauthoryear{Omidshafiei, Hennes, Garnelo, Wang, Recasens, Tarassov, Yang, Elie, Connor, Muller, et~al\mbox{.}}{Omidshafiei et~al\mbox{.}}{2022}]%
        {omidshafiei2022multiagent}
\bibfield{author}{\bibinfo{person}{Shayegan Omidshafiei}, \bibinfo{person}{Daniel Hennes}, \bibinfo{person}{Marta Garnelo}, \bibinfo{person}{Zhe Wang}, \bibinfo{person}{Adria Recasens}, \bibinfo{person}{Eugene Tarassov}, \bibinfo{person}{Yi Yang}, \bibinfo{person}{Romuald Elie}, \bibinfo{person}{Jerome~T Connor}, \bibinfo{person}{Paul Muller}, {et~al\mbox{.}}} \bibinfo{year}{2022}\natexlab{}.
\newblock \showarticletitle{Multiagent off-screen behavior prediction in football}.
\newblock \bibinfo{journal}{\emph{Scientific reports}} \bibinfo{volume}{12}, \bibinfo{number}{1} (\bibinfo{year}{2022}), \bibinfo{pages}{8638}.
\newblock


\bibitem[\protect\citeauthoryear{Pleuler}{Pleuler}{2024}]%
        {pleuler2024blog}
\bibfield{author}{\bibinfo{person}{Devin Pleuler}.} \bibinfo{year}{2024}\natexlab{}.
\newblock \bibinfo{title}{Unexpected Origins and the Fermi Paradox}.
\newblock \bibinfo{howpublished}{\url{https://www.centralwinger.com/p/unexpected-origins-and-the-fermi}}.
\newblock
\newblock
\shownote{Accessed: 12-02-2024.}


\bibitem[\protect\citeauthoryear{Radke, Brecht, and Radke}{Radke et~al\mbox{.}}{2022a}]%
        {radke2022identifying}
\bibfield{author}{\bibinfo{person}{David Radke}, \bibinfo{person}{Tim Brecht}, {and} \bibinfo{person}{Daniel Radke}.} \bibinfo{year}{2022}\natexlab{a}.
\newblock \showarticletitle{Identifying Completed Pass Types and Improving Passing Lane Models}. In \bibinfo{booktitle}{\emph{Link{\"o}ping Hockey Analytics Conference}}. \bibinfo{pages}{71--86}.
\newblock


\bibitem[\protect\citeauthoryear{Radke, Larson, and Brecht}{Radke et~al\mbox{.}}{2022b}]%
        {Radke2022Exploring}
\bibfield{author}{\bibinfo{person}{David Radke}, \bibinfo{person}{Kate Larson}, {and} \bibinfo{person}{Tim Brecht}.} \bibinfo{year}{2022}\natexlab{b}.
\newblock \showarticletitle{Exploring the Benefits of Teams in Multiagent Learning}. In \bibinfo{booktitle}{\emph{IJCAI}}.
\newblock


\bibitem[\protect\citeauthoryear{Radke and Orchard}{Radke and Orchard}{2023}]%
        {radke2023presenting}
\bibfield{author}{\bibinfo{person}{David Radke} {and} \bibinfo{person}{Alexi Orchard}.} \bibinfo{year}{2023}\natexlab{}.
\newblock \showarticletitle{Presenting multiagent challenges in team sports analytics}.
\newblock \bibinfo{journal}{\emph{AAMAS}} (\bibinfo{year}{2023}).
\newblock


\bibitem[\protect\citeauthoryear{Spearman}{Spearman}{2018}]%
        {spearman2018beyond}
\bibfield{author}{\bibinfo{person}{William Spearman}.} \bibinfo{year}{2018}\natexlab{}.
\newblock \showarticletitle{Beyond expected goals}. In \bibinfo{booktitle}{\emph{Proceedings of the 12th MIT sloan sports analytics conference}}. \bibinfo{pages}{1--17}.
\newblock


\bibitem[\protect\citeauthoryear{Spearman, Basye, Dick, Hotovy, and Pop}{Spearman et~al\mbox{.}}{2017}]%
        {spearman2017physics}
\bibfield{author}{\bibinfo{person}{William Spearman}, \bibinfo{person}{Austin Basye}, \bibinfo{person}{Greg Dick}, \bibinfo{person}{Ryan Hotovy}, {and} \bibinfo{person}{Paul Pop}.} \bibinfo{year}{2017}\natexlab{}.
\newblock \showarticletitle{Physics-based modeling of pass probabilities in soccer}. In \bibinfo{booktitle}{\emph{Proceeding of the 11th MIT Sloan Sports Analytics Conference}}, Vol.~\bibinfo{volume}{1}.
\newblock


\bibitem[\protect\citeauthoryear{Tuyls, Omidshafiei, Muller, Wang, Connor, Hennes, Graham, Spearman, Waskett, Steel, et~al\mbox{.}}{Tuyls et~al\mbox{.}}{2021}]%
        {tuyls2021game}
\bibfield{author}{\bibinfo{person}{Karl Tuyls}, \bibinfo{person}{Shayegan Omidshafiei}, \bibinfo{person}{Paul Muller}, \bibinfo{person}{Zhe Wang}, \bibinfo{person}{Jerome Connor}, \bibinfo{person}{Daniel Hennes}, \bibinfo{person}{Ian Graham}, \bibinfo{person}{William Spearman}, \bibinfo{person}{Tim Waskett}, \bibinfo{person}{Dafydd Steel}, {et~al\mbox{.}}} \bibinfo{year}{2021}\natexlab{}.
\newblock \showarticletitle{Game Plan: What {AI} can do for Football, and What Football can do for AI}.
\newblock \bibinfo{journal}{\emph{Journal of Artificial Intelligence Research}}  \bibinfo{volume}{71} (\bibinfo{year}{2021}), \bibinfo{pages}{41--88}.
\newblock


\bibitem[\protect\citeauthoryear{Wang, Veli{\v{c}}kovi{\'c}, Hennes, Toma{\v{s}}ev, Prince, Kaisers, Bachrach, Elie, Wenliang, Piccinini, et~al\mbox{.}}{Wang et~al\mbox{.}}{2024}]%
        {wang2024tacticai}
\bibfield{author}{\bibinfo{person}{Zhe Wang}, \bibinfo{person}{Petar Veli{\v{c}}kovi{\'c}}, \bibinfo{person}{Daniel Hennes}, \bibinfo{person}{Nenad Toma{\v{s}}ev}, \bibinfo{person}{Laurel Prince}, \bibinfo{person}{Michael Kaisers}, \bibinfo{person}{Yoram Bachrach}, \bibinfo{person}{Romuald Elie}, \bibinfo{person}{Li~Kevin Wenliang}, \bibinfo{person}{Federico Piccinini}, {et~al\mbox{.}}} \bibinfo{year}{2024}\natexlab{}.
\newblock \showarticletitle{TacticAI: an AI assistant for football tactics}.
\newblock \bibinfo{journal}{\emph{Nature communications}} \bibinfo{volume}{15}, \bibinfo{number}{1} (\bibinfo{year}{2024}), \bibinfo{pages}{1906}.
\newblock


\end{thebibliography}

% \input{demo_requirements}

%%%%%%%%%%%%%%%%%%%%%%%%%%%%%%%%%%%%%%%%%%%%%%%%%%%%%%%%%%%%%%%%%%%%%%%%

\end{document}